\documentclass[10pt,twocolumn,letterpaper]{article}

\usepackage{cvpr}
\usepackage{times}
\usepackage{epsfig}
\usepackage{graphicx}
\usepackage{amsmath}
\usepackage{amssymb}
\usepackage{authblk}
\usepackage{bm}
\usepackage{subcaption}
\usepackage{bbm}
\usepackage{algorithm}
\usepackage{algorithmic}

\newcommand{\rom}[1]{\lowercase\expandafter{\emph{\romannumeral #1}\relax}}

\usepackage[pagebackref=true,breaklinks=true,letterpaper=true,colorlinks,bookmarks=false]{hyperref}

\cvprfinalcopy 
\def\cvprPaperID{3040} 

\ifcvprfinal\fi

\begin{document}
\title{StoryGAN: A Sequential Conditional GAN for Story Visualization}
\author[1]{Yitong Li\thanks{This work was done while the first author was an intern at Microsoft Dynamics 365 AI Research.}}

\author[2]{Zhe Gan}
\author[4]{Yelong Shen}
\author[2]{Jingjing Liu}
\author[2]{Yu Cheng}
\author[5]{Yuexin Wu}
\author[1]{\\Lawrence Carin}
\author[1]{David Carlson}
\author[3]{Jianfeng Gao}
\affil[1]{Duke University, $^2$Microsoft Dynamics 365 AI Research, $^3$Microsoft Research} \affil[4]{Tencent AI Research, $^5$Carnegie Mellon University}
\renewcommand\Authands{ and }

\maketitle

\begin{abstract}
We propose a new task, called Story Visualization. Given a multi-sentence paragraph, the story is visualized by generating a sequence of images, one for each sentence. In contrast to video generation, story visualization focuses less on the continuity in generated images (frames), but more on the global consistency across dynamic scenes and characters -- a challenge that has not been addressed by any single-image or video generation methods. 
We therefore propose a new story-to-image-sequence generation model, StoryGAN, based on the sequential conditional GAN framework. Our model is unique in that it consists of a deep Context Encoder that dynamically tracks the story flow, and two discriminators at the story and image levels, to enhance the image quality and the consistency of the generated sequences.
To evaluate the model, we modified existing datasets to create the CLEVR-SV and Pororo-SV datasets. Empirically, StoryGAN outperforms state-of-the-art models in image quality, contextual consistency metrics, and human evaluation. 
\end{abstract}

\section{Introduction}\label{sec:intro}

\begin{figure}[htb]
\centering
\includegraphics[width=\columnwidth]{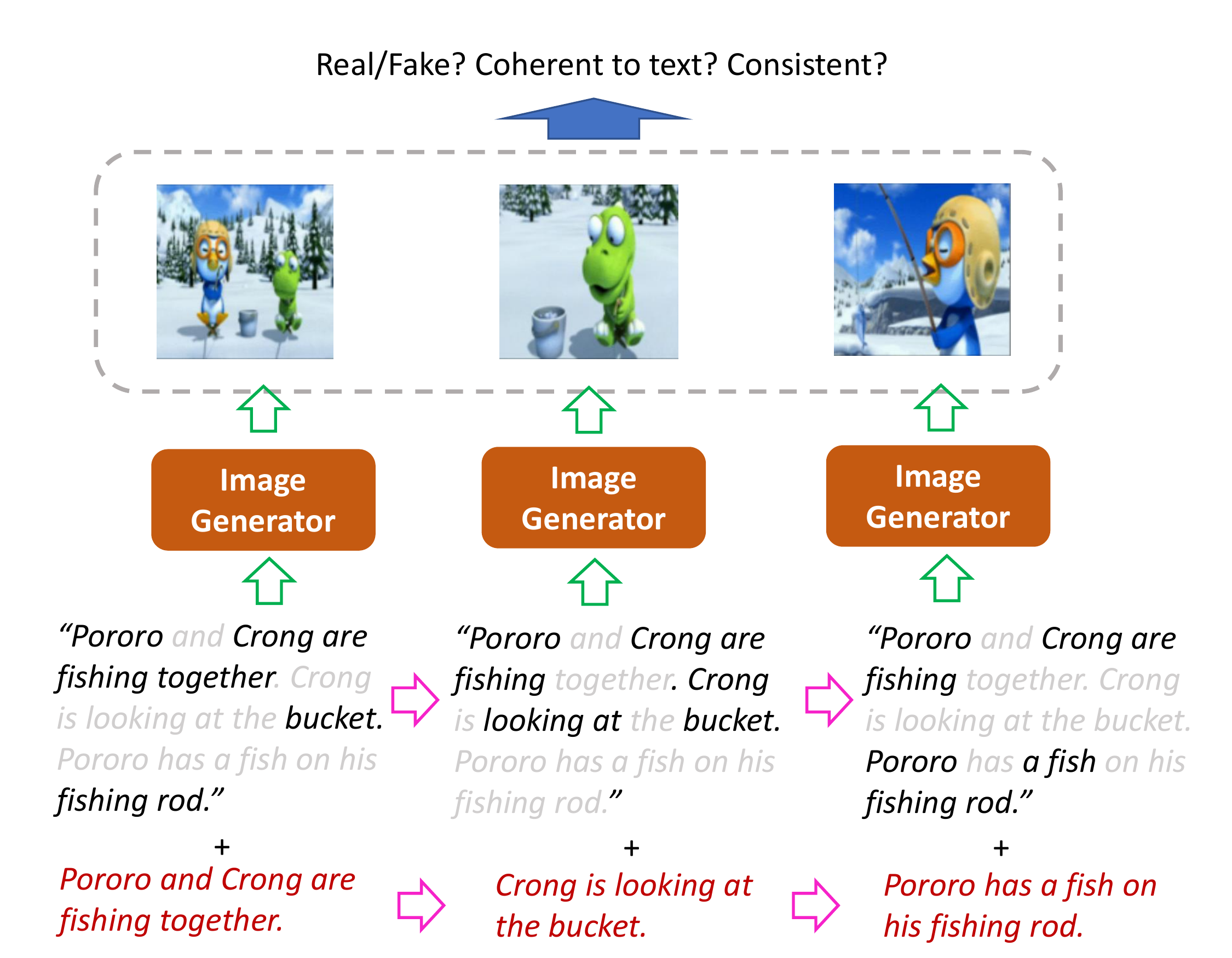}
\caption{\small The input story is ``Pororo and Crong are fishing together. Crong is looking at the bucket. Pororo has a fish on his fishing rod.'' Each sentence is visualized with one image. In this work, the image generation for each sentence is enriched with contextual information from the Context Encoder. Two discriminators at different levels guide the generation process.\vspace{-4mm}}
\label{fig:sv_sample}
\end{figure}

Learning to generate meaningful and coherent sequences of images from a natural language story is a challenging task, requiring understanding and reasoning on both natural language and images. We propose a new \emph{Story Visualization} task. Specifically, the goal is to generate a sequence of images to describe a story written in a multi-sentence paragraph, as shown in Figure~\ref{fig:sv_sample}. 

There are two main challenges in this task. First, the sequence of images must consistently and coherently depict the whole story.
This task is highly related to text-to-image generation~\cite{yan2016attribute2image,reed2016generative,isola2017image, zhang2017stackgan,xu2017attngan}, where an image is generated based on a short description. However, sequentially applying text-to-image methods to a story will not generate a coherent image sequence, failing on the story visualization task.
For instance, consider the  story ``A red metallic cylinder cube is at the center. Then add a green rubber cube at the right.'' The second sentence alone does not capture the entire scene.

The second challenge is how to display the logic of the storyline. Specifically, the appearance of objects and the layout in the background must evolve in a coherent way as the story progresses. This is similar to video generation. However, story visualization and video generation differ as: (\rom{1}) Video clips are \emph{continuous} with smooth motion transitions, so video generation models focus on extracting dynamic features to maintain realistic motions~\cite{vondrick2016generating,tulyakov2017mocogan}. In contrast, the goal of story visualization is to generate a sequence of key static frames that present correct story plots where motion features are less important. (\rom{2}) 
Video clips are often based on simple-sentence input and typically have a static background, while complex stories require the model to capture scene changes necessary for the plot line. In that sense, story visualization could also be viewed as a critical step towards real-world long-video generation by capturing sharp scene changes.
To tackle these challenges, we propose a StoryGAN framework, inspired by Generative Adversarial Networks (GANs)~\cite{goodfellow2014generative}, a two-player game between a generator and a discriminator. To take into account the contextual information in the sequence of sentence inputs, StoryGAN is designed as a sequential conditional GAN model. 

Given a multi-sentence paragraph (story), StoryGAN uses a recurrent neural network (RNN) to incorporate the previously generated images into the current sentence's image generation. 
Contextual information is extracted with our Context Encoder module, including a stack of a GRU cell and our newly proposed Text2Gist cell. The Context Encoder transforms the current sentence and a story encoding vector into a high-dimensional feature vector (Gist) for further image generation. As the story proceeds, the Gist is dynamically updated to reflect the change of objects and scenes in the story flow.
In the Text2Gist component, the sentence description is transformed into a filter and adapted to the story, so that we can optimize the mixing process by tweaking the filter. Similar ideas are also used in dynamic filtering~\cite{jia2016dynamic}, attention models~\cite{xu2017attngan} and meta-learning~\cite{rebuffi2018efficient}.

To ensure consistency across the sequence of generated images, we adopt a two-level GAN framework. We use an image-level discriminator to measure the relevance of a sentence and its generated image, and a story-level discriminator to measure the global coherence between the generated image sequence and the whole story. 

We created two datasets from the existing CLEVR~\cite{johnson2017clevr} and Pororo~\cite{kim2016pororoqa} datasets for our story visualization task, called CLEVR-SV and Pororo-SV, respectively. Empirically, StoryGAN more efficiently captures the full picture of the story and how it evolves, compared to existing baselines~\cite{zhang2017stackgan,li2018video}. Equipped with the deep Context Encoder module and the two-level discriminators, StoryGAN significantly outperforms previous state-of-the-art models, generating a sequence of higher quality images that are coherent with the story in both image quality and global consistency metrics, as well as human evaluation.

\section{Related Work}\label{sec:related_works}

Variational AutoEncoders (VAEs) ~\cite{kingma2013auto}, Generative Adversarial Nets (GANs)~\cite{goodfellow2014generative}, and flow-based generative models~\cite{dinh2014nice,dinh2016density}) have been widely applied to a wide range of generation tasks, including text-to-image generation, video generation, style transfer, and image editing. Story visualization falls into this broad categorization of generative tasks, but has several distinct aspects.   

Very relevant for the story visualization task is conditional text-to-image transformation~\cite{reed2016generative,isola2017image,zhu2017unpaired,yan2016attribute2image}, which can now generate high-resolution realistic images~\cite{zhang2017stackgan,xu2017attngan}. A key task in text-to-image generation is understanding longer and more complex input text.  For example, this has been explored in dialogue-to-image generation, where the input is a complete dialogue session rather than a single sentence ~\cite{sharma2018chatpainter}. Another related task is textual image editing, which edits an input image according to a textual editing query
~\cite{chen2017language,shetty2018adversarial,cheng2018sequential,el2018keep}. This task requires consistency between the original image and the output image.  Finally, there is the task of placing pre-specified images and objects in a picture from a text description ~\cite{kim2017codraw}.  This task also relates text to a consistent image, but does not require a full image-generation procedure.

A second closely related task to story visualization is video generation, especially that of text-to-video~\cite{li2018video,he2018probabilistic} or image-to-video generation~\cite{cai2017deep,tulyakov2017mocogan,vondrick2016generating}. 
Existing approaches only generate short video clips~\cite{he2018probabilistic,pmlr-v80-denton18a,hao2018controllable} without scene changes. The biggest challenge in video generation is how to ensure a smooth motion transition across successive video frames. A trajectory, skeleton or simple landmark is used in existing work, to help model the motion feature~\cite{hao2018controllable,zhao2018learning, wang2018every}. To this end, researchers disentangle dynamic and static features for motion and background, respectively~\cite{vondrick2016generating, li2018video,tulyakov2017mocogan,denton2017unsupervised}. In our modeling of story visualization, the whole story sets the static features and each input sentence encodes dynamic features. However, there are several differences: $(i)$ conditional video generation has only one input, while our task has sequential, evolving inputs; and $(ii)$ the motion in video clips is continuous, while images visualizing a story are discrete and often with different scene views.

There are also several other related tasks in the literature. For instance, story image retrieval from a pre-collected training set rather than image generation~\cite{ravi2018show}. Cartoon generation has been explored with a ``cut and paste'' technique~\cite{gupta2018imagine}. However, both of these techniques require large amounts of labeled training data. An inverse task to story visualization is visual storytelling, where the output is a paragraph describing a sequence of input images. Text generation models or reinforcement learning are often highlighted for visual storytelling~\cite{huang2016visual,liang2017recurrent,huang2018hierarchically}.

\section{StoryGAN}\label{sec:model}

\begin{figure*}[t]
\centering
\includegraphics[width=0.99\textwidth]{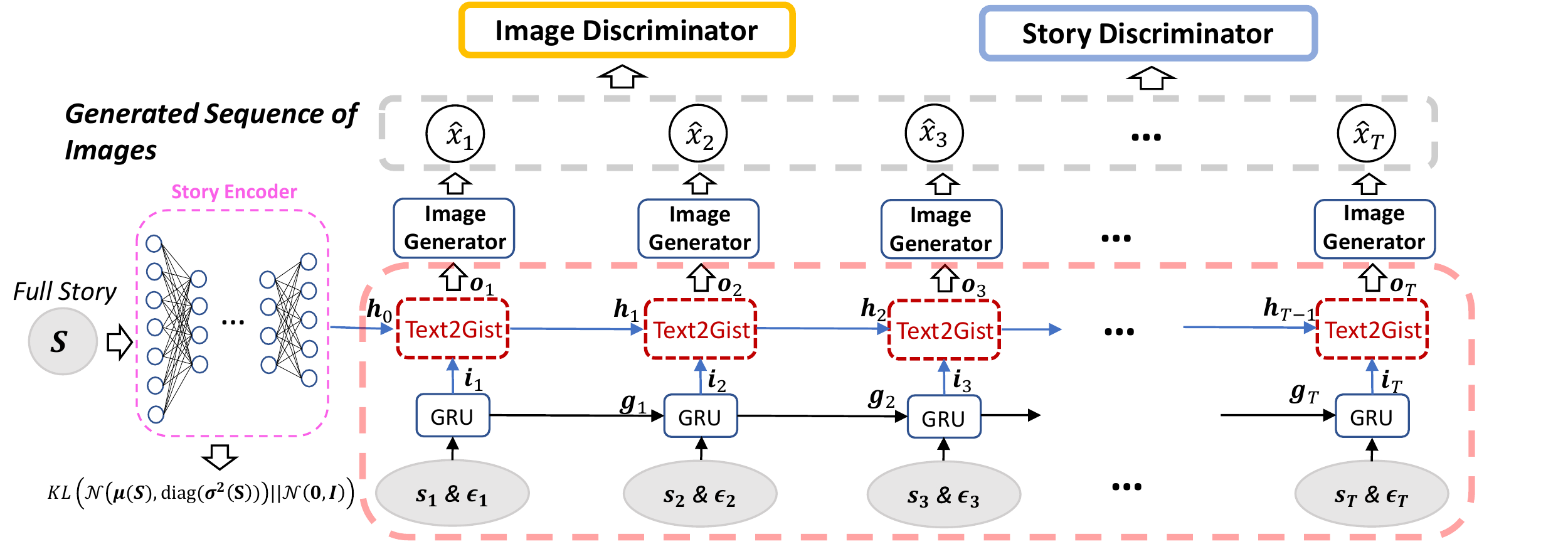}
\caption{\small The framework of StoryGAN. The variables in gray solid circles are the input story $\bm S$ and individual sentences $\bm s_1,\dots,\bm s_T$ with random noise $\bm \epsilon_1,\dots,\bm \epsilon_T$. The generator network contains the Story Encoder, Context Encoder and image generator. The proposed component Text2Gist is introduced in detail in Section~\ref{subsec:context_encoder}. There are two discriminators on top, which discriminate whether each image-sentence pair and each image-sequence-story pair are real or fake.\label{fig:framework}}

\end{figure*} 

StoryGAN is designed to create a sequence of images to describe an input story $S$. The story $S$ consists of a sequence of sentences $S=[\bm s_1, \bm s_2,\cdots,\bm s_T]$, where the length $T$ may vary. There is one generated image per sentence, denoted $\hat{\bm X} = [\hat{\bm x}_1, \hat{\bm x}_2, \cdots, \hat{\bm x}_T ]$, that are both locally (sentence-image) and globally (story-images) consistent. For training, ground truth images are denoted $\bm X = [\bm x_1, \bm x_2, \cdots, \bm x_T ]$. The image sequence is locally consistent if each image matches its corresponding sentence semantically. The image sequence is globally consistent if all the images globally hold together as coherent to the full story $S$ it visualizes. In our approach, each sentence in story $S$ has been encoded into an embedding vector using a pre-trained sentence encoder~\cite{cer2018universal}. 
With slight abuse of notation, each sentence is encoded via vector $\bm s_t \in \mathbb{R}^{128}$. In the following, we assume $\bm s_t$ and $\bm S$ are both encoded feature vectors instead of raw text.

The overall architecture of StoryGAN is presented in Figure~\ref{fig:framework}. It is implemented as a sequential GAN model, which consists of (\rom{1}) a Story Encoder that encodes $S$ into a low-dimensional vector ${\bm h}_0$; (\rom{2}) a two-layer recurrent neural network (RNN) based Context Encoder that encodes input sentence $s_t$ and its contextual information into a vector ${\bm o}_t$ (Gist) for each time point $t$; (\rom{3}) an image generator that generates image $\hat{\bm x}_t$ based on ${\bm o}_t$ for each time step $t$; and (\rom{4}) an image discriminator and a story discriminator that guide the image generation process so as to ensure the generated image sequence $\hat{\bm X}$ is locally and globally consistent, respectively.

\subsection{Story Encoder}\label{subsec:story_encoder}

The Story Encoder is given in the dotted pink box of Figure~\ref{fig:framework}. Following the conditioning mechanism in StackGAN~\cite{zhang2017stackgan}, the Story Encoder $E(\cdot)$ learns a stochastic mapping from story $S$ to an low-dimensional embedding vector ${\bm h}_0$. $\bm h_0$ encodes the whole story and it serves as the initial state of the hidden cell of the Context Encoder. Specifically, the Story Encoder samples an embedding vector ${\bm h}_0$ from 
a normal distribution ${\bm h}_0 \sim E(\bm S) = \mathcal{N}\left( \bm \mu(\bm S), \bm \Sigma(\bm S) \right)$, with $\bm \mu(\cdot)$ and $\bm \Sigma(\cdot)$ implemented as two neural networks. We restrict $\bm \Sigma(\bm S)=\text{diag}(\bm \sigma^2(\bm S))$ to a diagonal matrix for computational tractability. With the reparameterization trick, the encoded story $\bm h_0$ can be written as $\bm h_0 = \bm \mu(\bm S) + \bm \sigma^2(\bm S)^{\frac{1}{2}}\odot \bm \epsilon_S$, where $\bm \epsilon_S \sim \mathcal{N}(\bm 0, \bm I)$. Symbol $\odot$ represents elementwise multiplication, and the square root is also taken elementwise; $\bm \mu(\bm S)$ and $\bm \sigma^2(\bm S)$ are parameterized as Multi-Layer Perceptrons (MLPs) with a single hidden layer. Convolutional networks could also be used, depending on the structure of $\bm S$. The sampled $\bm h_0$ is provided to the RNN-based Context Encoder as the initial state vector.

By using stochastic sampling, the Story Encoder deals with the discontinuity problem in the original story space, thus not only leading to a compact, semantic representation of $S$ for story visualization, but also adding randomness to the generation process. The encoder's parameters are optimized jointly with the other modules of StoryGAN 
via back propagation. Therefore, to enforce the smoothness over the conditional manifold in latent semantic space and avoid collapsing to a single generative point rather than a distribution, we add the regularization term~\cite{zhang2017stackgan}, 
\begin{equation}\label{eq:story_encoder_KL_loss}
\mathcal{L}_{KL} = KL\left( \mathcal{N}\left(\bm \mu(\bm S), \text{diag}(\bm \sigma^2(\bm S)) \right)|| \mathcal{N}\left(\bm 0, \bm I \right)\right) ,
\end{equation}
which is the Kullback-Leibler (KL) divergence
between the learned distribution and the standard Gaussian distribution.


\subsection{Context Encoder}\label{subsec:context_encoder}

Video generation is closely related to story visualization, and it typically assumes a static background with smooth motion transitions, requiring a disjoint embedding of static and dynamic features~\cite{vondrick2016generating,huang2016visual,tulyakov2017mocogan}.  In story visualization, the challenge differs in that the characters, motion, and background often change from image to image, as illustrated in Figure~\ref{fig:sv_sample}. This requires addressing two problems: (\rom{1}) how to update the contextual information to effectively capture background changes; and (\rom{2}) how to combine new inputs and random noise when generating each image, to visualize the change of characters, which may shift dramatically. 

We address these issues by proposing a deep RNN based Context Encoder to capture contextual information during sequential image generation, shown in the red box in Figure~\ref{fig:framework}. The context can be defined as any related information in the story that is useful for the current generation.
The deep RNN consists of two hidden layers. The lower layer is implemented using standard GRU cells and the upper layer using the proposed Text2Gist cells, which are a variant to GRU cells and are detailed below. At time step $t$, the GRU layer takes as input the concatenation of the sentence $\bm s_t$ and isometric Gaussian noise $\bm \epsilon_t$, and outputs the vector ${\bm i}_t$. The Text2Gist cell combines the GRU's output ${\bm i}_t$ with the story context ${\bm h}_t$ (initialized by Story Encoder) to generate ${\bm o}_t$ that encodes all necessary information for generating an image at time $t$. ${\bm h}_t$ is updated by the Text2Gist cell to reflect the change of potential context information.

Let $\bm g_t$ and $\bm h_t$ denote the hidden vectors of the GRU and Text2Gist cells, respectively. The Context Encoder works in two steps to generate its output: 
\begin{align}
\bm i_t, \bm g_t &= \text{GRU} (\bm s_t, \bm \epsilon_t, \bm g_{t-1}), \\  
\bm o_t, \bm h_t &= \text{Text2Gist} (\bm i_t, \bm h_{t-1}). 
\end{align}
We call $\bm o_t$ the \emph{``Gist'' vector} since it combines all the global and local context information, from $\bm h_{t-1}$ and $\bm i_t$ respectively, at time step $t$ ($i.e.$, it captures the ``gist'' of the information).
The Story Encoder initializes $\bm h_0$, while $\bm g_0$ is randomly sampled from an isometric Gaussian distribution. 

We next give the underlying updates of Text2Gist.
Given $\bm h_{t-1}$ and $\bm i_t$ at time step $t$, Text2Gist generates a hidden vector $\bm h_t$ and an output vector $\bm o_t$ as follows:
\begin{align}
\bm z_t & = \sigma_z \left( \bm W_z \bm i_t + \bm U_t \bm h_{t-1} + \bm b_z \right) \label{eq:zt_Text2Gist}, \\ 
\bm r_t & = \sigma_r \left( \bm W_r \bm i_t + \bm U_r \bm h_{t-1} + \bm b_r \right) \label{eq:rt_Text2Gist},\\ 
\bm h_t & =  (\bm 1 - \bm z_t) \odot \bm h_{t-1} \nonumber \\
& + \bm z_t \odot \sigma_h \left( \bm W_h \bm i_t + \bm U_h (\bm r_t \odot \bm h_{t-1}) + \bm b_h  \right) \label{eq:ht_Text2Gist}, \\
\bm o_t & =  \text{Filter} (\bm i_t) * \bm h_t, \label{eq:ot_Text2Gist}
\end{align}
where $\bm z_t$ and $\bm r_t$ are the outputs from the update and reset gates, respectively. 
The update gate decides how much information from the previous step should be kept, and the reset gate determines what to forget from $\bm h_{t-1}$. 
$\sigma_z$, $\sigma_r$ and $\sigma_h$ are sigmoid non-linearity functions. 
{
In contrast to standard GRU cells, 
output $\bm o_t$ is the convolution between $\text{Filter} (\bm i_t)$ and $\bm h_t$. 
The filter $\bm i_t$ is learned to adapt to $\bm h_t$. Specifically, $\text{Filter}(\cdot)$ transforms vector $\bm i_t$ to a multi-channel filter of size $C_{out} \times 1 \times 1 \times \text{len}(\bm h_t)$ using a neural network, where $C_{out}$ is the number of output channels. Since $\bm h_t$ is a vector, this filter is used as a 1D filter as in a standard convolutional layer.
}

The convolution operator in Eq.~\eqref{eq:ot_Text2Gist} infuses the global contextual information from $\bm h_t$ and local information from $\bm i_t$. $\bm o_t$ is the output of the Text2Gist cell at time step $t$. 
Since $\bm i_t$ encodes information from $\bm s_t$ and $\bm h_t$ from $\bm S$, which reflects the whole picture of the story, the convolutional operation in Eq.~\eqref{eq:ot_Text2Gist} can be seen as helping $\bm s_t$ to pick out the important part from the story in the process of generation. Empirically, we find that Text2Gist is more effective than traditional RNNs for story visualization.


\subsection{Discriminators}\label{subsec:discriminators}


StoryGAN uses two discriminators, an image and a story discriminator, to ensure the local and global consistency of the story visualization, respectively. The image discriminator measures whether the generated image ${\hat x}_t$ matches the sentence $s_t$ given its initial context information encoded in ${\bm h}_{0}$.  It does this by comparing the generated triplet $\{\bm s_t,{\bm h}_0,\hat{\bm  x}_t\}$ to the real triplet $\{\bm s_t,{\bm h}_0,\bm {x}_t\}$.
In contrast to prior work on text-to-image generation~\cite{zhang2017stackgan,reed2016generative}, the same sentence can have a significantly different generated image depending on the context, so it is important to give the encoded context information to the discriminator as well.
For example, consider the example given in Section~\ref{sec:intro}, ``A red metallic cylinder cube is at the center. Then add a green rubber cube at the right of it.'' The second image will vary wildly without the context ($i.e.$, the first sentence).

\begin{figure}[htb]
\centering
\includegraphics[width=.5\textwidth]{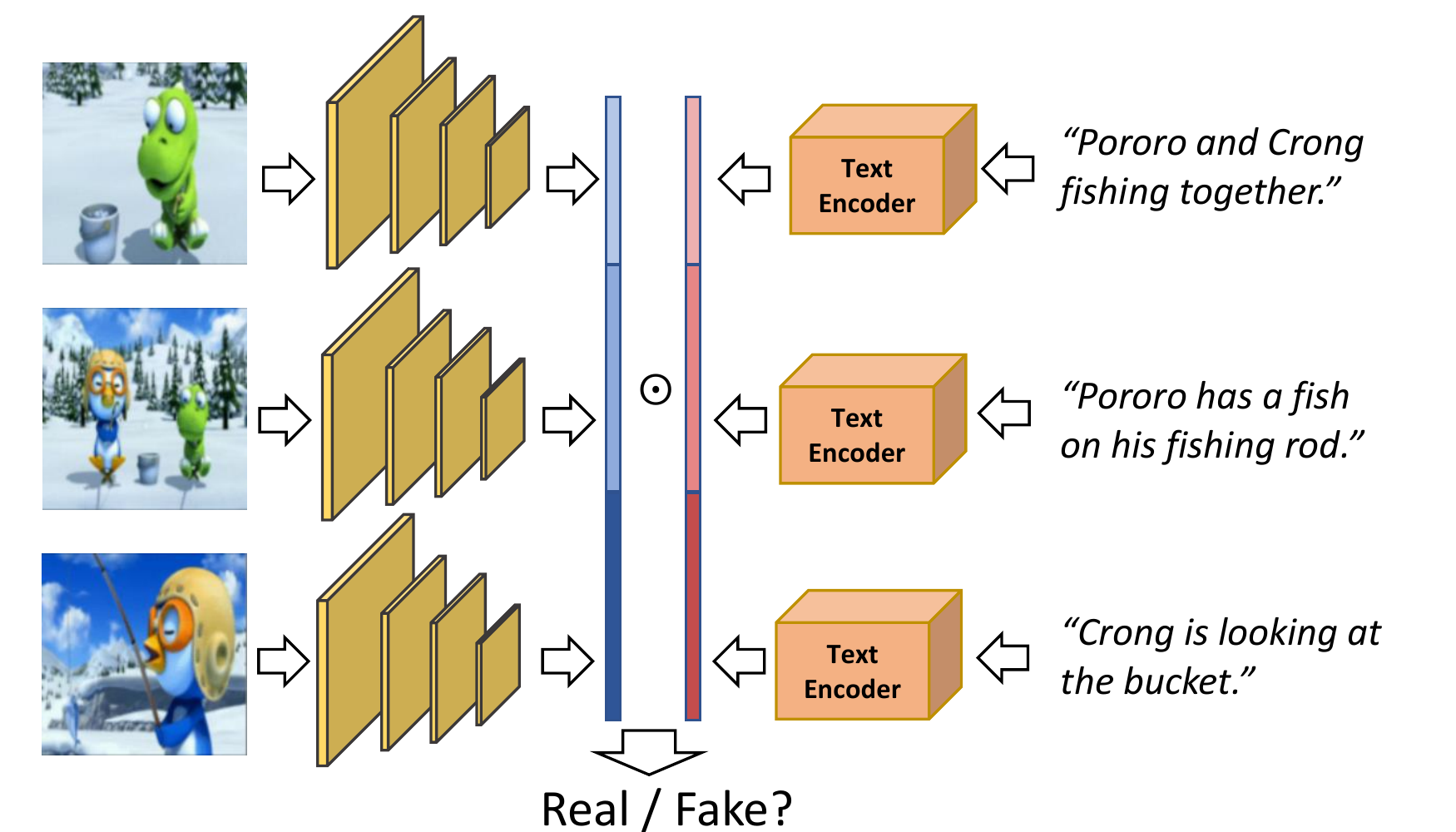}
\caption{\small Structure of the story discriminator. The feature vectors of the images/sentences in the story are concatenated. The product of image and text features are input to a fully connected layer with sigmoid non-linearity to predict whether it is a fake or real story pair.}
\label{fig:story_discriminator}
\end{figure} 

The story discriminator helps enforce the global consistency of the generated image sequence given story $S$. It differs from the discriminators used for video generation, which often use 3D convolution~\cite{vondrick2016generating, tulyakov2017mocogan,li2018video} to smooth the changes between frames.  
The overall architecture of the story discriminator is illustrated in Figure~\ref{fig:story_discriminator}. The left part is an image encoder, which encodes an image sequence into a sequence of feature vectors $E_{img}(\bm X) = \left[E_{img}(\bm x_1), \cdots, E_{img}(\bm x_T) \right]$, where $\bm X$ are either real or generated images (which are denoted by $\Hat{\bm X}$). These vectors are concatenated into a single vector, shown 
as the blue rectangular in Fig.~\ref{fig:story_discriminator}. 
Similarly, the right part is a text encoder, which encodes the multi-sentence story $S$ into a sequence of feature vectors $E_{txt}(\bm S) = [E_{txt}(\bm s_1), \cdots, E_{txt}(\bm s_T)]$. Likewise, these are concatenated into one big vector, shown as the red rectangle in Fig.~\ref{fig:story_discriminator}.
The image encoder is implemented as a deep convolutional network and the text encoder as a multi-layer perceptron. Both output a same dimensional vector. 

The global consistency score is computed as
\begin{equation}\label{eq:d_loss}
D_S = \sigma \left(\bm w^\intercal \left(E_{img}(\bm X) \odot E_{txt}(\bm S) \right)+ b \right) ,
\end{equation}
where $\odot$ is element-wise product. The weights $\bm w$ and bias $\bm b$ are learned in the output layer.  $\sigma$ is a sigmoid function that normalizes the score to a value in $[0,1]$. By pairing each sentence and image, the story discriminator can consider both local matching and global consistency jointly. 

Both image and story discriminators are trained on positive and negative pairs. The latter are generated by replacing the image (sequence) in the positive pairs with generated ones.


\subsection{Algorithm Outlines}\label{subsec:alg_outline}
Let $\bm \theta$, $\bm \psi_I$, and $\bm \psi_S$ denote the parameters of the whole generator $G(\cdot; \bm \theta)$, the image discriminator, and the story discriminator, respectively. The objective function for StoryGAN is
\begin{equation}\label{eq:final_objective}
\min_{\bm \theta} \max_{\bm \psi_I, \bm \psi_S} \alpha \mathcal{L}_{Image} + \beta \mathcal{L}_{Story} + \mathcal{L}_{KL},
\end{equation}
where $\alpha$ and $\beta$ balance the three loss terms. $\mathcal{L}_{KL}$ is the regularization term of the Story Encoder defined in~\eqref{eq:story_encoder_KL_loss}. $\mathcal{L}_{Image}$ and $\mathcal{L}_{Story}$ are defined as
\begin{align}\label{eq:final_loss}
\mathcal{L}_{Image} & = \textstyle{\sum_{t=1}^T} (\mathbb{E}_{(\bm x_t, \bm s_t)}\left[\log D_{I} (\bm x_t, \bm s_t, \bm h_0; \bm \psi_I)\right] \nonumber \\
& + \mathbb{E}_{(\bm \epsilon_t, \bm s_t)} \left[ \log (1- D_I(G(\bm \epsilon_t, \bm s_t;\bm  \theta), \bm s_t, \bm h_0; \bm \psi_I) ) \right] ) \\
\mathcal{L}_{Story} & = \mathbb{E}_{(\bm X, \bm S)}\left[\log D_{S} (\bm X, \bm S; \bm \psi_S)\right] \nonumber \\
& + \mathbb{E}_{(\bm \epsilon, \bm S)} \left[ \log (1- D_S(\left[G(\bm \epsilon_t, \bm s_t; \bm \theta)\right]_{t=1}^T), \bm S; \bm \psi_S) )\right] .
\end{align}
$D_I(\cdot; \bm \psi_I)$ and $D_S(\cdot;\bm \psi_S)$ are the image and story discriminator, parameterized by $\bm \psi_I$ and $\bm \psi_S$, respectively.



The pseudo-code for training StoryGAN is given in Algorithm~\ref{alg:storygan}. The parameters of the story and image discriminators, $\bm \psi_I$ and $\bm \psi_S$, are updated in two separate \textbf{for} loops, respectively, while the parameters of the image generator $\bm \theta$ are updated in both loops. 
The initial hidden state of the Text2Gist layer is the encoded story feature vector $\bm h_0$ produced by the Story Encoder\footnote{Code is available at \url{https://github.com/yitong91/StoryGAN} }. The detailed configuration of the network is provided in Appendix~\ref{appen:network_structure}.

\begin{algorithm}[h]
\caption{Training Procedure of StoryGAN}\label{alg:storygan}
\begin{algorithmic}
\STATE \textbf{Input}: Encoded sentence vectors $\bm S_n=[\bm s_{n1},\bm s_{n2}, \cdots, \bm s_{nT}]$ and corresponding images $\bm X_n=[\bm x_{n1}, \cdots, \bm x_{nT}]$ for $n=1,\cdots,N$.
\STATE \textbf{Output}: Generator parameters $\bm \theta$ and discriminator parameters $\bm \psi_I$ and $\bm \psi_S$.
\\\hrulefill
\FOR{$iter = 1$ to $max\_iter$}
  \FOR{$iter_I$ = 1 to $k_I$}
    \STATE Sample a mini-batch of story-sentence pairs $\{(\bm s_t, \bm S, \bm x_t) \}$ from the training set. 
    \STATE Compute $\bm h_0$ as the initialization of the Text2Gist layer and the KL regularization term as Eq.~\eqref{eq:story_encoder_KL_loss}.
    \STATE Generate a single output image ${\hat x}$. 
    \STATE Update $\bm \psi_I$ and $\bm \theta$.
  \ENDFOR
  \FOR{$iter_S$ = 1 to $k_S$}
    \STATE Sample a mini-batch of story-image pair $\{(\bm S, \bm X) \}$ from training set.
    \STATE Compute $\bm h_0$ and update $\bm h_t$ at each time step $t$
    \STATE Generate image sequence $\hat{\bm X}$. 
    \STATE Update $\bm \psi_S$ and $\bm \theta$.
  \ENDFOR
\ENDFOR
\end{algorithmic}
\end{algorithm}

In our experiments, we use Adam~\cite{kingma2014adam} for parameter updates. We also find that using different mini-batch sizes for image and story discriminators may accelerate training convergence, and that it is beneficial to update generator and discriminator in different time steps in one epoch. 

\section{Experiments}\label{sec:experiment}
In this section, we evaluate the StoryGAN model on one toy and one cartoon dataset. To the best of our knowledge, there is no existing work on our proposed story visualization task. The closest alternative for story visualization is conditional video generation~\cite{li2018video}, where the story is treated as single input and a video is generated in lieu of the sequence of  images. However, we empirically found that the video generation result is too blurry and not comparable to StoryGAN. Thus, our comparisons are mainly to ablated versions of our proposed model. For a fair comparison, all models use the same structure of the image generator, Context Encoder and discriminators when applicable. The compared baseline models are:

\textbf{ImageGAN:} ImageGAN follows the work in~\cite{reed2016generative,zhang2017stackgan} and does not use the story discriminator, story encoder and Context Encoder. Each image is generated independently. However, for a reasonable comparison, we concatenate $\bm s_t$, encoded story $\bm S$ and a noise term as input. Otherwise, the model fails on the task. This is the simplest version of StoryGAN.

\begin{figure}[htb]
\centering
\includegraphics[width=0.5\textwidth]{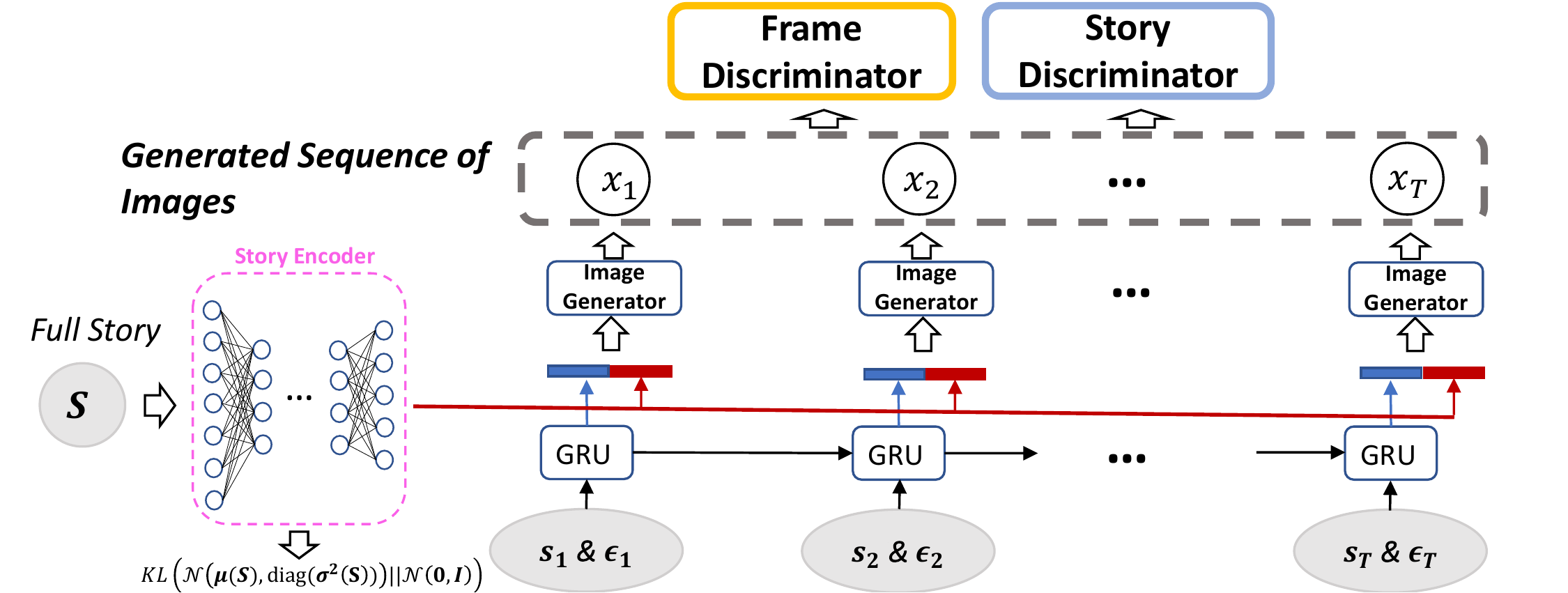}
\caption{\small The framework of the baseline model SVC, where the story and individual sentence are concatenated to form the input.}
\label{fig:framework_svc}
\end{figure}

\textbf{SVC:} In ``Story Visualization by Concatenation'' (SVC), the Text2Gist cell in StoryGAN is replaced by simple concatenation of the encoded story and description feature vectors~\cite{tulyakov2017mocogan}. Compared to ImageGAN, SVC includes the additional story discriminator, and is visualized in Figure~\ref{fig:framework_svc}.

\textbf{SVFN}: In ``Story Visualization by Filter Network'' (SVFN), the concatenation in SVC is replaced by a filter network. Sentence $\bm s_t$ is transformed into a filter and convolved with the encoded story. Specifically, the image generator input is $\bm o_t = \text{Filter}(\bm i_t) * \bm h_0$ instead of Eq.~\ref{eq:ot_Text2Gist}.

\subsection{CLEVR-SV Dataset}\label{subsec:exp-clevr}
\begin{figure*}[htb]
\centering
\includegraphics[width=1.0\textwidth]{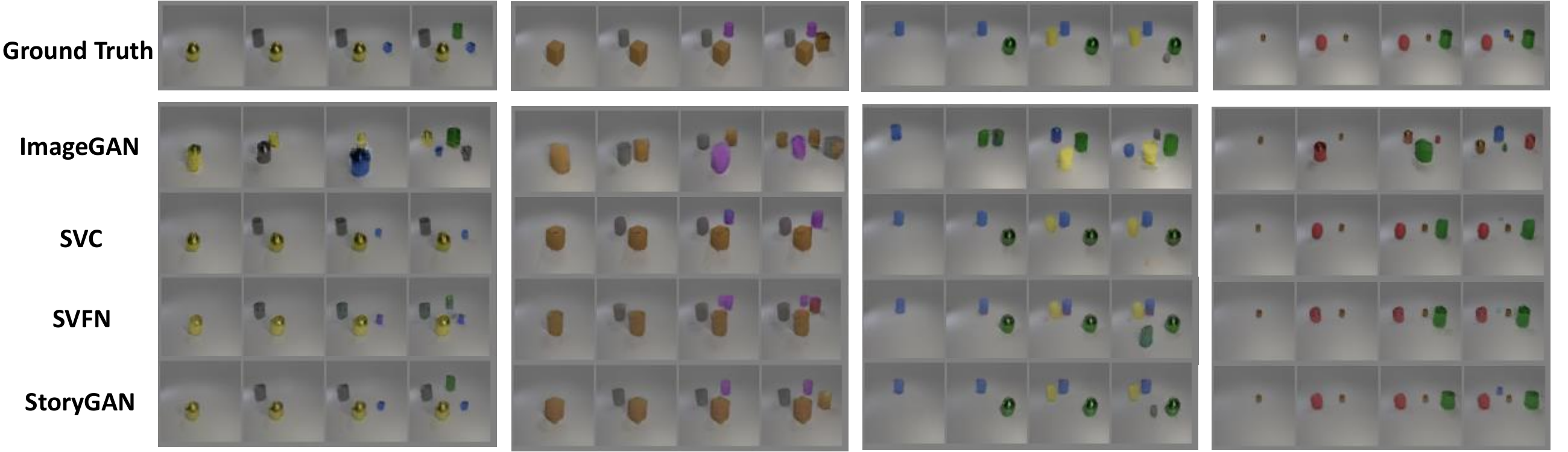}
\caption{Comparison among different methods on CLEVR-SV dataset. }
\label{fig:clevr_samples}
\vspace{-4mm}
\end{figure*} 
The CLEVR~\cite{johnson2017clevr} dataset was originally used for visual question answering. We modified this data for story visualization by generating images from randomly assigned layouts of the object (examples in the top row of Figure~\ref{fig:clevr_samples}). We named this dataset CLEVR-SV to distinguish it from the existing CLEVR dataset. Specifically, four rules were used to construct the CLEVR-SV: $(i)$ The maximum number of objects in one story is limited to four. $(ii)$ Objects are made of metallic/rubber with eight different colors and two different sizes. $(iii)$ The object shape can be cylinder, cube or sphere. $(iv)$ The object is added one at a time, resulting in a four-image sequence per story. We generated $10,000$ image sequences for training and $3,000$ for testing. For our task, the story is the layout descriptions of objects.

The input $\bm s_t$ is the current object's attribute and the relative position given by two real numbers indicating its coordinates. For instance, the first image of the left column of Fig.~\ref{fig:clevr_samples} is generated from ``yellow, large, metal, sphere, (-2.1, 2.4).'' The following objects are described in the same way. Given the description, the generated objects' appearance should have 
little variation from the ground truth and their relative positions should be similar. 

Figure~\ref{fig:clevr_samples} gives the results comparison. ImageGAN~\cite{reed2016generative} fails to keep the consistency of the `story' and it mixes up the attributes when the number of objects increases. SVC solves this consistency problem by including the story discriminator and GRU cell at the bottom, as the third row of Figure~\ref{fig:clevr_samples} has consistent objects in the image sequence. However, SVC generates an implausible forth image in the sequence. We hypothesize that using simple vector concatenation cannot effectively balance the importance of the current description with the whole story. SVFN can alleviate this problem to some extent, but not completely. In contrast, StoryGAN generates more feasible images than the competitors. We attribute the performance improvement to three components: $(i)$ Text2Gist cell tracks the progress of story; $(ii)$ story and image discriminators keep the consistency of objects in the generation process; $(iii)$ using the Story Encoder to initialize the Text2Gist cell gives better result on first generated image. Greater empirical evidence for this final point appears in the cartoon dataset in Section~\ref{subsec:exp_cartoon}.

In order to further validate the StoryGAN model, we designed a task to evaluate whether the 
model can generate consistent images by changing the first sentence description. Specifically, we randomly replaced the first object's description while keeping the other three the same during generation, which we visualize in Supplemental Figure~\ref{fig:clevr_change_samples_compare} in Appendix~\ref{appen:clevr}.  This comparison shows that only StoryGAN can keep the story consistency by correctly utilizing the attributes of the first object in later frames, as discussed above.  In Supplemental Figure~\ref{fig:clevr_change_samples}, we give additional examples on changing the initial attributes only using StoryGAN.  Regardless of the initial attribute, StoryGAN is consistent between frames.

\begin{table}[htb]
  \centering
  \begin{tabular}{l|c|c|c|c} 
  \hline
  & ImageGAN~\cite{reed2016generative} & SVC & SVFN & StoryGAN \\
  \hline
  SSIM & 0.596 & 0.641 & 0.654 & 0.672\\
  \hline
  \end{tabular}
  \caption{SSIM comparison on CLEVR-SV dataset.} \label{tab:clevr_SSIM}
\end{table}

We also compare the Structural Similarity Index (SSIM) score between the generated images and ground truth~\cite{hore2010image}. SSIM was originally used to measure the recovery result from distorted images. Here, it is used to determine whether the generated images are aligned with the input description. Table~\ref{tab:clevr_SSIM} gives the SSIM metric for each method on the test set. Note that though this is a generative task, using SSIM to measure the structure similarity is reasonable because there is little variation given the description. In this task, StoryGAN significantly outperforms the other baselines.

\subsection{Cartoon Dataset}\label{subsec:exp_cartoon}
\begin{figure*}[htb]
\centering
\includegraphics[width=0.9\textwidth]{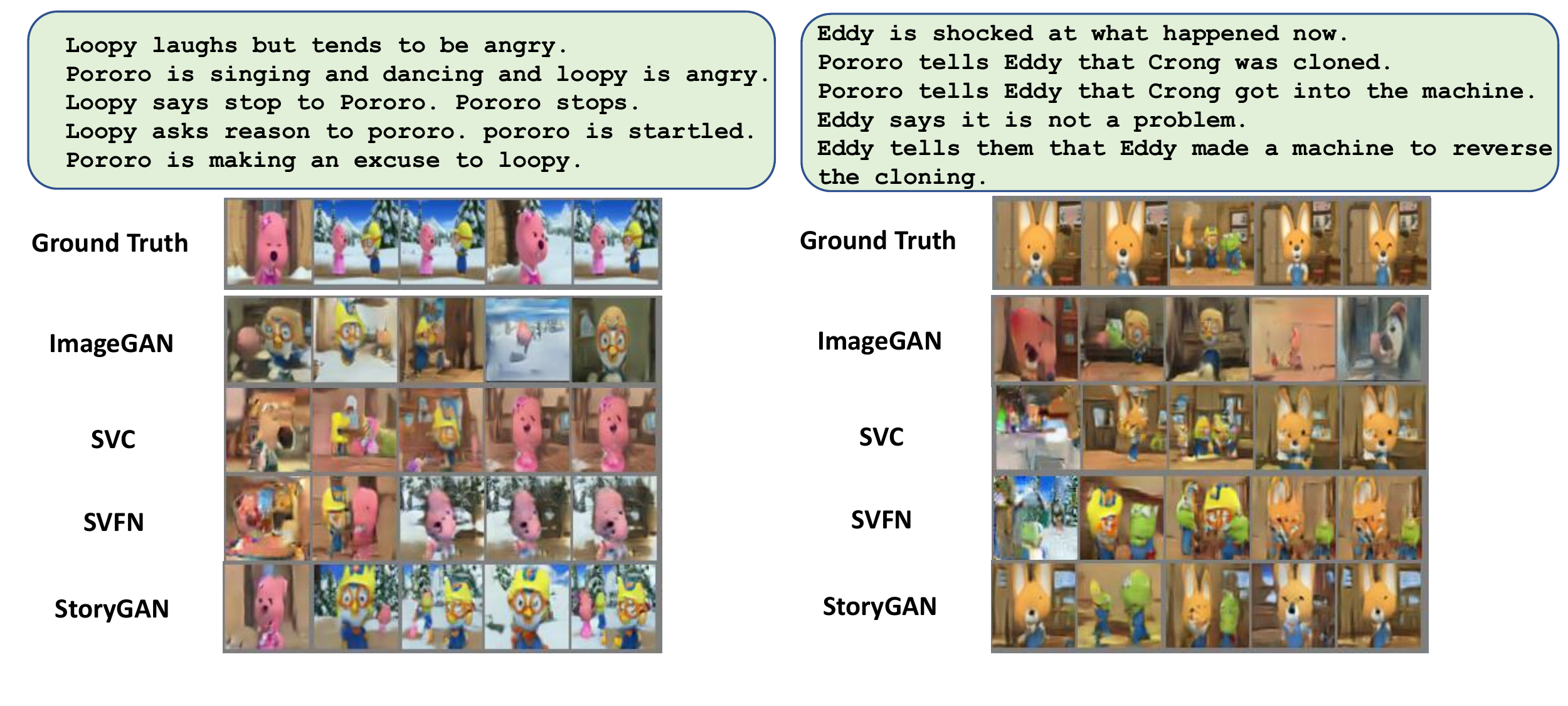}
\vspace{-5mm}
\caption{ Two generated samples on the Pororo-SV dataset. }
\label{fig:pororo_stage1_result}
\end{figure*}
The Pororo dataset~\cite{kim2016pororoqa} was originally used for video question answering, where each one second video clip is associated with more than one manually written description. About $40$ video clips forms a complete story. Each story has several QA pairs. In total, the Pororo dataset contains $16$K clips of one second videos about $13$ distinct characters. The manually written description has an average length of $13.6$ words that describes what is happening and which characters are in each video clip. These $16$K video clips are sorted into $408$ movie stories~\cite{kim2016pororoqa}.

We modified the Pororo dataset to fit story visualization task by considering the description for each video clip as the story's text input. For each video clip, we randomly pick out one frame (sampling rate is 30Hz) during training as the real image sample. Five continuous images form a single story. Finally, we end up with $15,336$ description-story pairs, where $13,000$ pairs are used as training, the remaining $2,336$ pairs for testing. We call this dataset Pororo-SV to differ it from the original Pororo QA dataset~\cite{kim2016pororoqa}.

The text encoder uses universal encoding~\cite{cer2018universal} with fixed pre-trained parameters.  Training a new text encoder empirically gave little performance gain. Two visualized stories from the competing methods are given in Figure~\ref{fig:pororo_stage1_result}. The text input is given on the top. ImageGAN does not generate consistent image sequences; for instance, the generated images switch from indoors to outdoors randomly.  Additionally, the characters' appearance is inconsistent in the sequence of images (e.g. Pororo's hat). SVC and SVFN can improve the consistency to some extent, but their limitations can be seen in the unsatisfactory first images. In contrast, StoryGAN's  first image has a much higher quality than other baselines because of the use of the Story Encoder to initialize the recurrent cell. This shows the advantage of using the output of the Story Encoder as first hidden state over random initialization.

\begin{figure}[htb]
\includegraphics[width=0.5\textwidth]{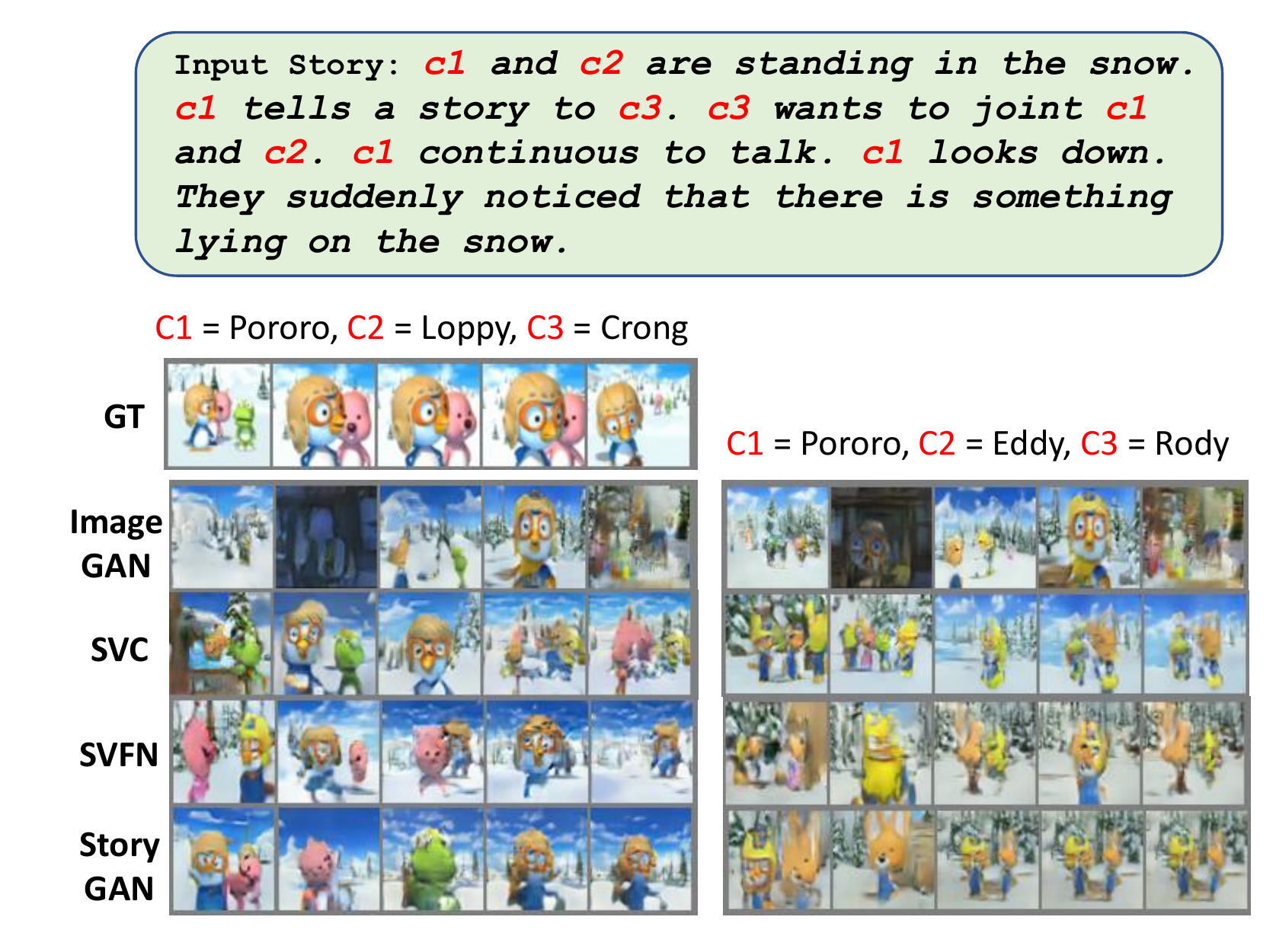}
\caption{ \small Generation result by changing character names in the same story.  The story template is given at the top, with the character names $c1$, $c2$ and $c3$ in the two instances of the story, each one shown in a column.}
\label{fig:pororo_change_samples}
\vspace{-4mm}
\end{figure}

To explore how different models represent the story, we ran experiments where only the character names in the story were changed, shown in Figure~\ref{fig:pororo_change_samples}. Visually, StoryGAN outperforms the other baselines on the image quality and consistency.

Further, we perform two distinct quantitative tasks. The first is to determine whether the generation is able to capture the relevant characters in the story. The nine most common characters are selected from the dataset. Their names and pictures are provided in Supplemental Figure~\ref{fig:clevr_change_samples} in Appendix~\ref{appen:pororo}. Next, a character image classifier is trained on real images from training set and applied on both real and generated images from the test set. We compare the classification accuracy (only exact matches across all characters counts as correct) of each image/story pair as an indicator of whether the generation is coherent to the story description. The classifier's performance on the test set is 86\%, which is considered an upper bound for the task.  Note that there is peculiarity in the labels, as the human-generated labels can sometimes include characters not shown in the frame. Furthermore, the classifier is trained on real images, not the generated images, so the domain gap between real and generated images may harm the performance. However, these issues should hurt all algorithms equally and it is a fair comparison.  From the results below, it is clear that StoryGAN has increased character consistency compared to the baseline models.

\begin{table}[htb]
  \centering
   \resizebox{1.0\columnwidth}{!}{
  \begin{tabular}{l|c|c|c|c|c} 
  \hline
  & Upper Bound & ImageGAN~\cite{reed2016generative} & SVC & SVFN & StoryGAN \\
  \hline
  Acc. & 0.86 & 0.23 & 0.21 & 0.24 & 0.27\\
  \hline
  \end{tabular} }
  \caption{\small Character classification accuracy (exact match ratio) comparison on Pororo-SV dataset.  The upper bound is the classifier accuracy on the real images associated with the stories.} \label{tab:pororo_acc}
\end{table}
\vspace{-6mm}
\paragraph{Human Evaluation} Automatic metrics cannot fully evaluate the performance of StoryGAN. Therefore, we performed both pairwise and ranking-based human evaluation studies on Amazon Mechanical Turk on Pororo-SV. For both tasks, we use 170 generated image sequences sampled from the test set, each assigned to 5 workers to reduce human variance. The order of the options within each assignment is shuffled to make a fair comparison. 

We first performed a pairwise comparison between StoryGAN and ImageGAN. For each input story, the worker is presented with two generated image sequences and asked to make decisions from the three aspects: visual quality\footnote{The generated images look visually appealing, rather than blurry and difficult to understand.}, consistency\footnote{The generated images are consistent with each other, have a common topic hidden behind, and naturally forms a story, rather than looking like 5 independent images.}, and relevance\footnote{The generated image sequence accurately reflects the input story and covers the main characters mentioned in the story.}. Results are summarized in Table~\ref{tbl:human_pair}. The standard error on these estimates is small, demonstrating that StoryGAN drastically outperformed ImageGAN on this task. 

We next performed ranking-based human evaluation.  For each input story, the worker is asked to rank images generated from the four compared models on their overall quality. Results are summarized in Table~\ref{tbl:human_rank}. StoryGAN achieves the highest average rank, while ImageGAN performs the worst. There is little uncertainty in these estimates, so we are confident that humans prefer StoryGAN on average.
\begin{table}[t!]
	\centering
	\caption{\small Results of pairwise human evaluation. The $\pm$ denotes standard error on the metrics.}
	\label{tbl:human_pair}
	\small
	\begin{tabular}{c|ccc}
		\hline
		& \multicolumn{3}{c}{StoryGAN vs ImageGAN}  \\
		\hline
		Choice (\%) & StoryGAN & ImageGAN & Tie  \\
		\hline
		Visual Quality & 74.17 \tiny{$\pm$1.38} & 18.60 \tiny{$\pm$1.38} & 7.23  \\ 
		Consistence & 79.15 \tiny{$\pm$1.27} & 15.28 \tiny{$\pm$1.27}& 5.57 \\ 
		Relevance & 78.08 \tiny{$\pm$1.34}& 17.65\tiny{$\pm$1.34} & 4.27  \\
		\hline
	\end{tabular}
	\vspace{-2mm}
\end{table}

\begin{table}[t!]
	\centering
	\caption{\small Results of ranking-based human evaluation.  The $\pm$ denotes standard error on the metrics. }
    \vspace{-2mm}
	\label{tbl:human_rank}
	\small
	\begin{tabular}{c|cccc}
		\hline
		Method & ImageGAN & SVC & SVFN & StoryGAN  \\
		\hline
		Rank & 2.91\tiny{$\pm$0.05} & 2.42\tiny{$\pm$0.04} & 2.77\tiny{$\pm$0.04} & 1.94\tiny{$\pm$0.05} \\ 
		\hline
	\end{tabular}
\end{table}
\section{Conclusion}\label{sec:convlusion}
We studied the story visualization task as a sequential conditional generation problem. The proposed StoryGAN model deals with the task by jointly considering the current input sentence with the contextual information. This is achieved by the proposed Text2Gist component in the Context Encoder. From the ablation test, the two-level discriminator and the recurrent structure on the inputs helps ensure the consistency across the generated images and the story to be visualized, while the Context Encoder efficiently provides the image generator with both local and global conditional information. Both quantitative and human evaluation studies show that StoryGAN improves the generation compared to the baseline models.  As image generators improve, the story visualization's quality will improve also. 

\small{
\bibliographystyle{ieee}
\bibliography{ref} }
\clearpage
\appendix
\section{Network Configuration}\label{appen:network_structure}
This section gives the network structure used in StoryGAN. In the following, `CONV' means the 2D convolutional layer, which is configured by output channel number `C', kernel size `K', step size `S' and padding size `P'. `LINEAR' is fully connected layer, with input and output dimensions given in the parenthesis. Note that the Filter Network is contained in the Text2Gist cell, which transforms $\bm i_t$ to a filter. This is introduced in detail in section~\ref{subsec:context_encoder}.
\begin{table}[htb]
	\centering
	\caption{\small Network Structure used in StoryGAN. * This layer combines the conditional input and the encoded images. }
    \vspace{-2mm}
	\label{tbl:clevr_layer}
	\small
	\begin{tabular}{c|l}
		\hline
	       Layer & Story Encoder\\
	       \hline
	       1 & LINEAR-(128 $\times$ T, 128), BN, RELU \\
	       \hline
	       Layer & Context Encoder \\
	       \hline
	       1 & LINEAR-(NOISEDIM + TEXTDIM, 128), BN, RELU \\
	       2 & GRU-(128, 128) \\
	       3 & Text2Gist-(128, 128) \\
	       \hline
	       Layer & Filter Network \\
	       \hline
	       1 & LINEAR-(128, 1024), BN, TANH\\
	       2 & RESHAPE(16, 1, 1, 64) \\
	       \hline
	       Layer & Image Generator \\
	       \hline
	       1 & CONV-(C512, K3, S1, P1), BN, RELU \\
	       2 & UPSAMPLE-(2,2) \\
	       3 & CONV-(C256, K3, S1, P1), BN, RELU \\
	       4 & UPSAMPLE-(2,2) \\
	       5 & CONV-(C128, K3, S1, P1), BN, RELU \\
	       6 & UPSAMPLE-(2,2) \\
	       7 & CONV-(C64, K3, S1, P1), BN, RELU \\
	       8 & UPSAMPLE-(2,2) \\
	       9 & CONV-(C3, K3, S1, P1), BN, TANH \\
	       \hline 
	       Layer & Image Discriminator \\
	       \hline
	       1 & CONV-(C64, K4, S2, P1), BN, LEAKY RELU  \\
	       2 & CONV-(C128, K4, S2, P1), BN, LEAKY RELU \\
	       3 & CONV-(C256, K4, S2, P1), BN, LEAKY RELU \\
	       4 & CONV-(C512, K4, S2, P1),BN, LEAKY RELU \\
	       5* & CONV-(C512, K3, S1, P1), BN, LEAKY RELU \\
	       6 & CONV-(C1, K4, S4, P0), SIGMOID \\
	       \hline
	       Layer & Story Discriminator (Image Encoder) \\
	       \hline
	       1 & CONV-(C64, K4, S2, P1), BN, LEAKY RELU  \\
	       2 & CONV-(C128, K4, S2, P1), BN, LEAKY RELU \\
	       3 & CONV-(C256, K4, S2, P1), BN, LEAKY RELU \\
	       4 & CONV-(C512, K4, S2, P1),BN, LEAKY RELU \\
	       5 & CONV-(C32, K4, S2, P1),BN, CONCAT \\
	       6 & RESHAPE-(1, 32 $\times$ 4 $\times$ T  ) \\
	       \hline
	       Layer & Story Discriminator (Text Encoder) \\
	       \hline
	       1 & LINEAR-(128 $\times$ T, 32 $\times$ 4 $\times$ T ), BN \\
		\hline
	\end{tabular}
	\vspace{-4mm}
\end{table}

\section{More Examples of CLEVR-SV Dataset}\label{appen:clevr}
\begin{figure}[htb]
\centering
\hspace{-5mm}
\includegraphics[width=0.45\textwidth]{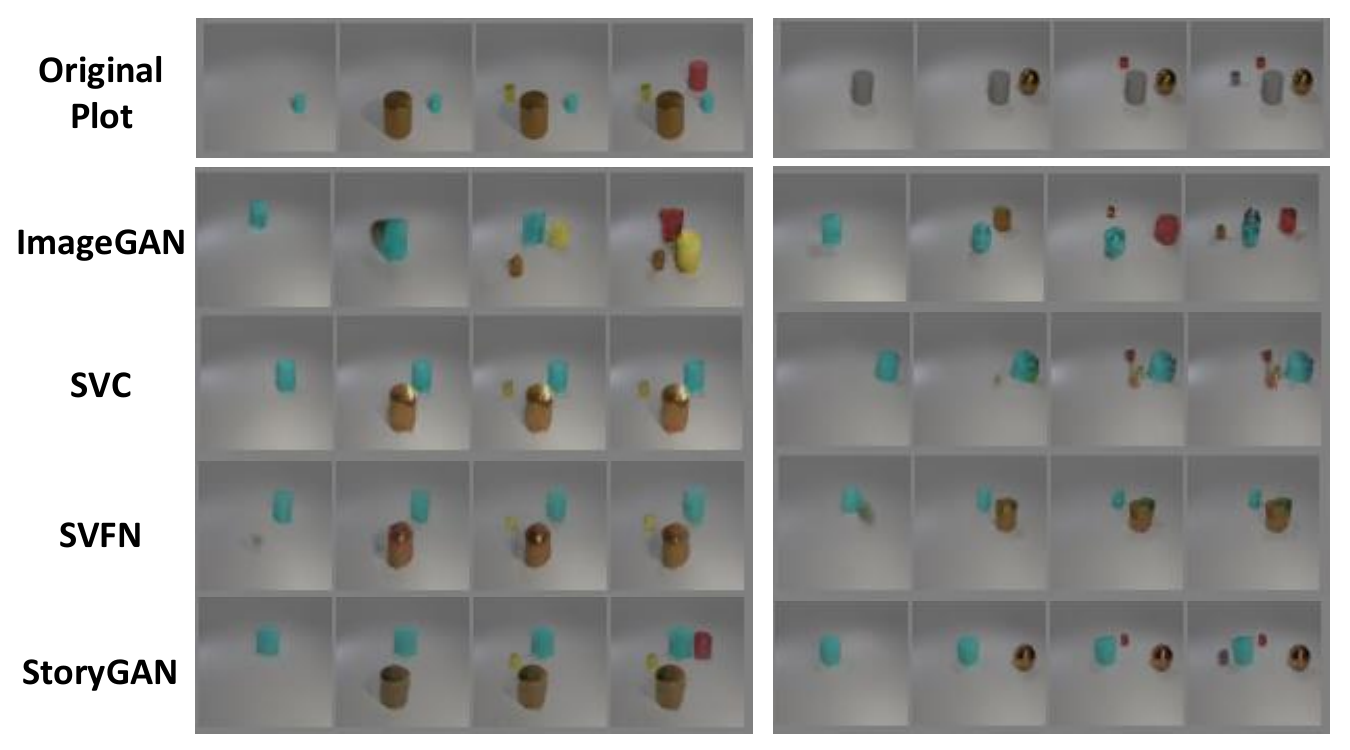}
\caption{\small Method comparison on a task where the original story description is changed in the first sentence.  Specifically, the first sentence is now ``Large, Rubber, Cyan, Cylinder, at (-0.46, -0.36).'' Each column corresponds to one layout of the following three objects. The first row is the original image that will be modified.  Note that only StoryGAN keeps the story consistency among the compared methods.}
\label{fig:clevr_change_samples_compare}
\end{figure}

\begin{figure}[htb]
\centering
\hspace{-5mm}
\includegraphics[width=0.45\textwidth]{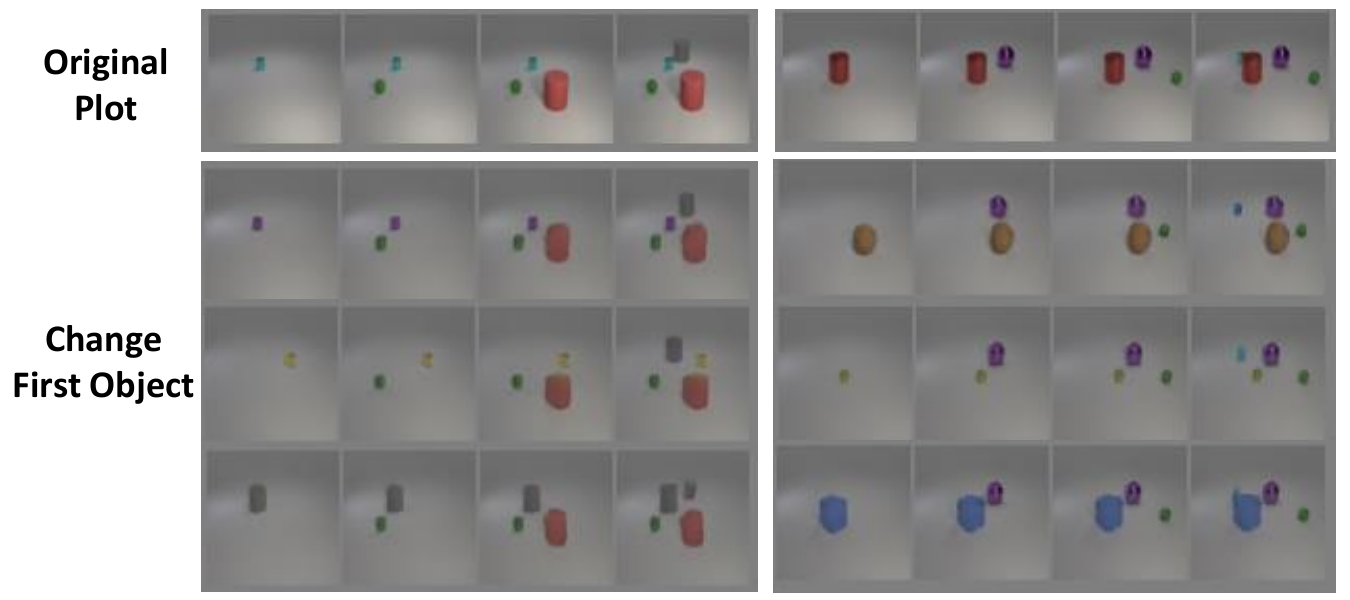}
\caption{\small An additional example using the same idea as Figure \ref{fig:clevr_change_samples_compare}. The top row gives two initial setups. The next three rows correspond to StoryGAN generations with different first sentences. For the left column, the attributes of the first object are:`Small, Metal, Cyan, Cylinder, at (-2.00, 0.02)' (original), `Small, Metal, Purple, Cylinder, at (-1.15, -0.26)', `Small, Metal, Yellow, Cylinder, at  (0.35, 2.00)' and `Large, Rubber, Gray, Cylinder, at (-1.77, -0.07)', respectively. For the right column, the attributes of the first object are: `Large, Metal, Red, Sphere, at (0.52, 0.56)' (original), `Large, Rubber, Brown, Sphere, at (-1.54, 0.85) ', `Small, Metal, Yellow, Cylinder, at  (-0.85, 2.29) ' and `Large, Rubber, Blue, Cube, at (0.15, -0.19) ', respectively. Again, we omit the attribute input of the second, third and forth objects to save the space.  Note that regardless of the initial description, StoryGAN effectively captures the story consistency. }
\label{fig:clevr_change_samples}
\end{figure}

Here we perform the test by using the same attributes of the first object. We test if the models can keep the first object consistent through the following generations. Figure~\ref{fig:clevr_change_samples_compare} compares the results from different models. Figure~\ref{fig:clevr_change_samples} gives more samples using StoryGAN.

\clearpage 
\section{Significance Test on Pororo-SV Dataset}\label{appen:pororo}
We perform pairwise t-test on the human evaluated ranking test. As we can see, StoryGAN is statistically significant over other baseline models.
\begin{table}[htb]
	\centering
	\caption{\small p-value on the human evaluated ranking test.}
    \vspace{-2mm}
	\label{tbl:human_rank_pvalue}
	\small
	\begin{tabular}{c|cccc}
		\hline
		Method & ImageGAN & SVC & SVFN & StoryGAN  \\
		\hline
		ImageGAN & 1.0 & 5e-13 & 0.04 & 1e-40\\
		SVC & 5e-13 & 1.0 & 1e-8 & 4e-14\\
		SVFN & 0.04 & 1e-8 &1.0 & 3e-36\\
		StoryGAN & 1e-40 & 4e-14 & 3e-36 & 1.0\\
		\hline
	\end{tabular}
	\vspace{-4mm}
\end{table}

\section{Characters Photo and More Examples of Pororo-SV Dataset}\label{appen:pororo}
\begin{figure}[htb]
\centering
\hspace{-5mm}
\includegraphics[width=0.49\textwidth]{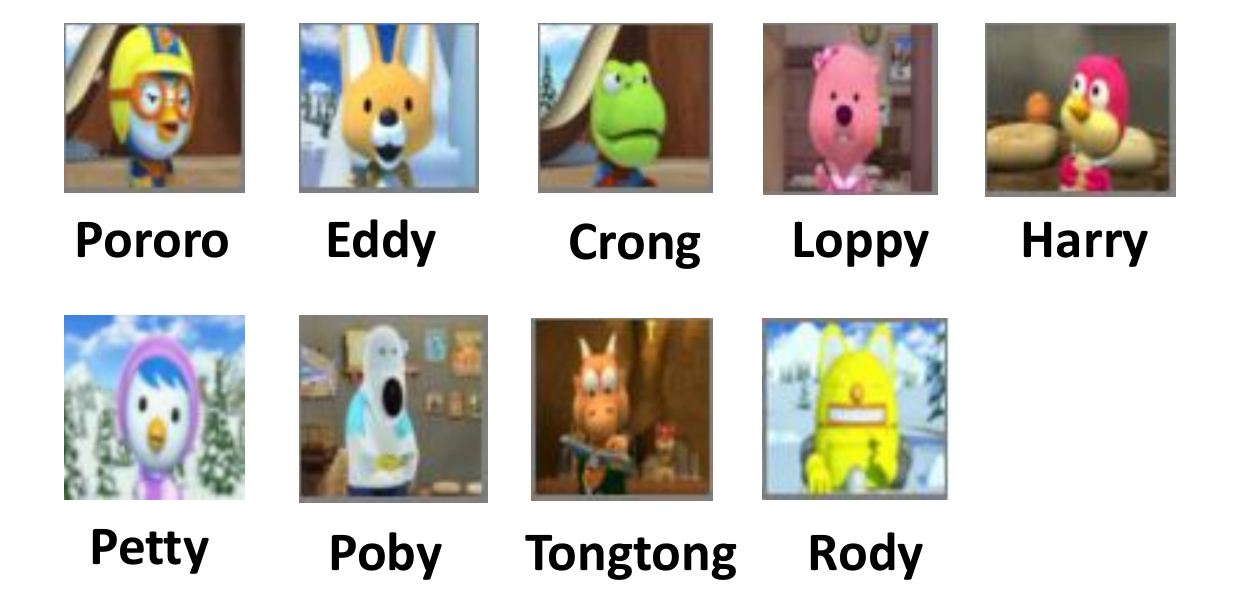}
\caption{\small Main character names and corresponding photos from the dataset.}
\label{fig:character_sample}
\end{figure}

For the classification accuracy compared in Table~\ref{tab:pororo_acc}, nine characters are selected: `Pororo', `Crong', `Eddy', `Poby', `Loopy', `Petty', `Harry', `Rody' and `Tongtong'. Profile pictures of them are given in Figure~\ref{fig:character_sample}.

\begin{figure}[htb]
\centering
\hspace{-5mm}
\includegraphics[width=0.45\textwidth]{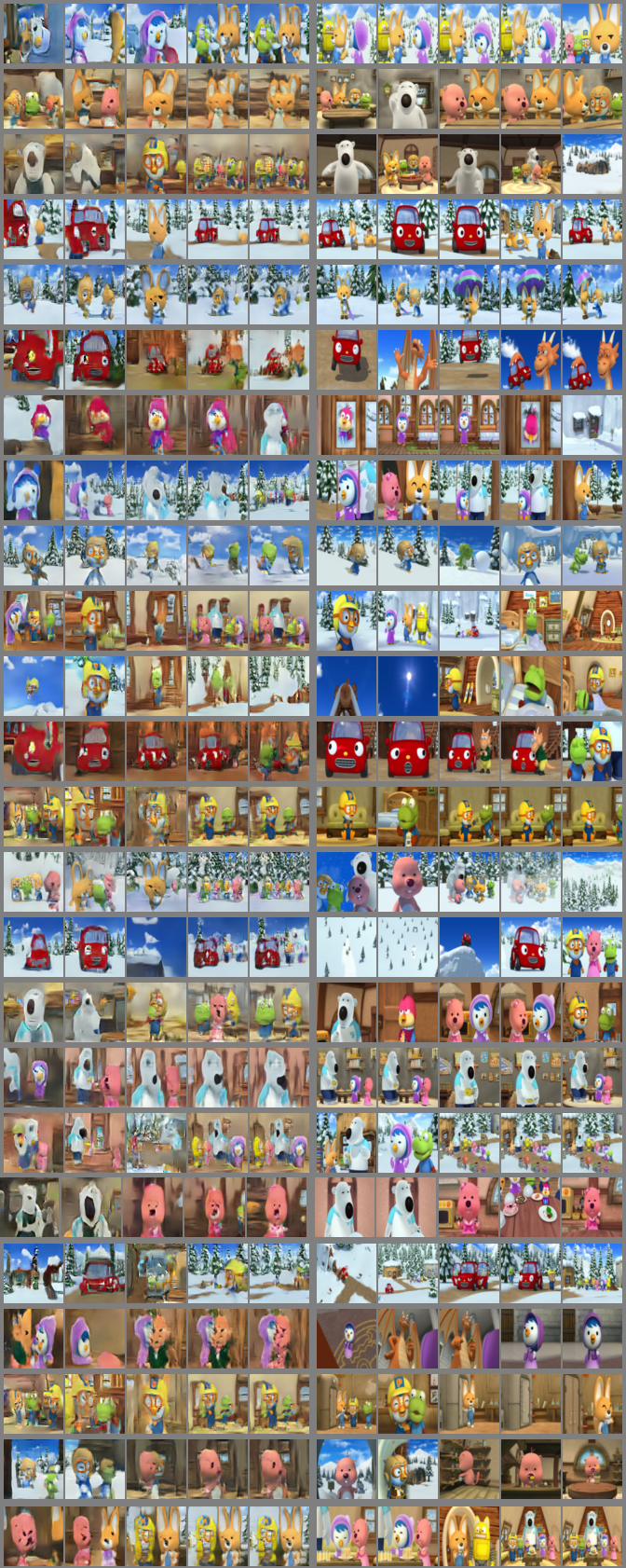}
\caption{\small More samples on Pororo-SV test set. For simplicity, we give the ground truth story images instead of the raw story text. The left five columns are generated images. The right five columns are ground truth. Note that there is no need for the generation to exactly match the ground truth. Those  samples with similar (but not exactly the same) images are caused by repeated input sentences.}
\label{fig:pororo_appen}
\end{figure}

Then, more generated samples on Pororo-SV dataset are given in Figure~\ref{fig:pororo_appen}.

\end{document}